\documentclass[runningheads]{llncs}
\usepackage{graphicx}
\usepackage{amsmath,amssymb} 
\usepackage{color}
\usepackage[width=122mm,left=12mm,paperwidth=146mm,height=193mm,top=12mm,paperheight=217mm]{geometry}

\usepackage[titletoc]{appendix}

\begin{document}
\pagestyle{headings}
\mainmatter

\title{ActiveStereoNet: End-to-End Self-Supervised Learning for Active Stereo Systems} 

\titlerunning{ActiveStereoNet}

\authorrunning{Y. Zhang et al.}

\author{Yinda Zhang$^{1,2}$, Sameh Khamis$^1$, Christoph Rhemann$^1$, Julien Valentin$^1$, Adarsh Kowdle$^1$, Vladimir Tankovich$^1$, Michael Schoenberg$^1$, \\ Shahram Izadi$^1$, Thomas Funkhouser$^{1,2}$, Sean Fanello$^1$   }


\institute{$^1$Google Inc., $^2$Princeton University 
}

\maketitle
\begin{abstract}
In this paper we present ActiveStereoNet, the first deep learning solution for active stereo systems. 
Due to the lack of ground truth, our method is fully self-supervised, yet it produces precise depth with a subpixel precision of $1/30th$ of a pixel; it does not suffer from the common over-smoothing issues; it preserves the edges; and it explicitly handles occlusions.
We introduce a novel reconstruction loss that is more robust to noise and texture-less patches, and is invariant to illumination changes. The proposed loss is optimized using a window-based cost aggregation with an adaptive support weight scheme. This cost aggregation is edge-preserving and smooths the loss function, which is key to allow the network to reach compelling results. Finally we show how the task of predicting invalid regions, such as occlusions, can be trained end-to-end without ground-truth. This component is crucial to reduce blur and particularly improves predictions along depth discontinuities. Extensive quantitatively and qualitatively evaluations on real and synthetic data demonstrate state of the art results in many challenging scenes.

\keywords{Active Stereo, Depth Estimation, Self-supervised Learning, Neural Network, Occlusion Handling, Deep Learning}
\end{abstract}


\section{Introduction}

Depth sensors are revolutionizing computer vision by providing additional 3D information for many hard problems, such as non-rigid reconstruction \cite{dou16,dou17}, action recognition \cite{fanello13,fanello2013one} and parametric tracking \cite{taylor16,taylor17} . Although there are many types of depth sensor technologies, they all have significant limitations. Time of flight systems suffer from motion artifacts and multi-path interference \cite{bhandari14b,bhandari14,naik15}.   Structured light is vulnerable to ambient illumination and multi-device interference \cite{fanello2017ultrastereo,hyperdepth}.  Passive stereo struggles in texture-less regions, where expensive global optimization techniques are required - especially in traditional non-learning based methods.

\begin{figure}[t]
    \centering
      \includegraphics[width=\columnwidth]{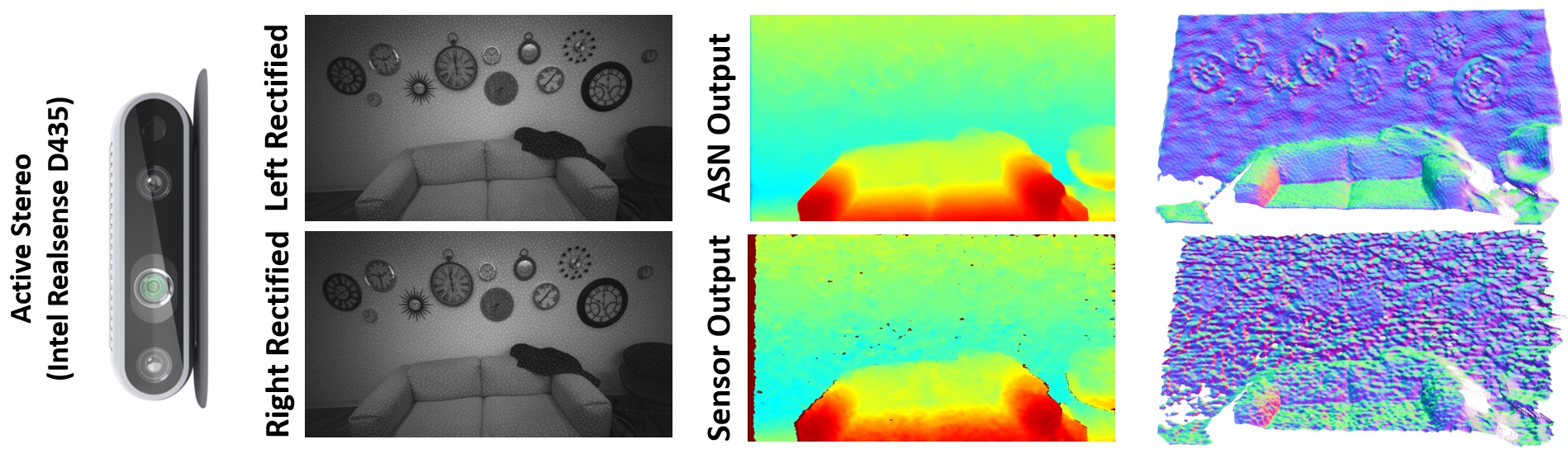}
    \caption{ActiveStereoNet (ASN) produces smooth, detailed, quantization free results using a pair of rectified IR images acquired with an Intel Realsense D435 camera. In particular, notice how the jacket is almost indiscernible using the sensor output, and in contrast, how it is clearly observable in our results.}
 \label{fig:teaser}
 \end{figure}

Active stereo offers a potential solution: an infrared stereo camera pair is used, with a pseudorandom pattern projectively texturing the scene via a patterned IR light source. (Fig. \ref{fig:teaser}). 
With a proper selection of sensing wavelength, the camera pair captures a combination of active illumination and passive light, improving quality above that of structured light while providing a robust solution in both indoor and outdoor scenarios.
Although this technology was introduced decades ago \cite{nishihara1984prism}, it has only recently become available in commercial products (e.g., Intel R200 and D400 family \cite{inteld435}).  As a result, there is relatively little prior work targeted specifically at inferring depths from active stereo images, and large scale training data with ground truth is not available yet.  

Several challenges must be addressed in an active stereo system.  Some are common to all stereo problems -- for example, it must avoid matching occluded pixels, which causes oversmoothing, edge fattening, and/or flying pixels near contour edges.  However, other problems are specific to active stereo -- for example, it must process very high-resolution images to match the high-frequency patterns produced by the projector; it must avoid the many local minima arising from alternative alignments of these high frequency patterns; and it must compensate for luminance differences between projected patterns on nearby and distant surfaces.  Additionally, of course, it cannot be trained with supervision from a large active stereo dataset with ground truth depths, since none is available.

This paper proposes the first end-to-end deep learning approach for active stereo that is trained fully self-supervised. It extends recent work on self-supervised passive stereo \cite{zhong2017self} to address problems encountered in active stereo.  
First, we propose a new reconstruction loss based on local contrast normalization (LCN) that removes low frequency components from passive IR and re-calibrates the strength of the active pattern locally to account for fading of active stereo patterns with distance. Second, we propose a window-based loss aggregation with adaptive weights for each pixel to increase its discriminability and reduce the effect of local minima in the stereo cost function.  Finally, we detect occluded pixels in the images and omit them from loss computations. These new aspects of the algorithm provide significant benefits to the convergence during training and improve depth accuracy at test time.  Extensive experiments demonstrate that our network trained with these insights outperforms previous work on active stereo and alternatives in ablation studies across a wide range of experiments.
 

\section{Related Work}

Depth sensing is a classic problem with a long history of prior work. Among the \textbf{active sensors}, Time of Flight (TOF), such as Kinect V2, emits a modulated light source and uses multiple observations of the same scene (usually 3-9) to predict a single depth map. The main issues with this technology are artifacts due to motion and multipath interference \cite{bhandari14b,bhandari14,naik15}.
Structure light (SL) is a viable alternative, but it requires a known projected pattern and is vulnerable to multi-device inference \cite{fanello2017ultrastereo,hyperdepth}. Neither approach is robust in outdoor conditions under strong illumination.

\textbf{Passive stereo} provides an alternative approach \cite{scharstein2002taxonomy,hamzah2016literature}.  Traditional methods utilize hand-crafted schemes to find reliable local correspondences \cite{bleyer2011patchmatch,yoon2005locally,hosni2013fast,bleyer2008simple,hirschmuller2008stereo} and global optimization algorithms to exploit context when matching \cite{besse2014pmbp,felzenszwalb2006efficient,klaus2006segment,kolmogorov2001computing}. Recent methods address these problems with deep learning.  Siamese networks are trained to extract patch-wise features and/or predict matching costs \cite{luo2016efficient,zbontar2016stereo,zagoruyko2015learning,zbontar2015computing}.   More recently, 
end-to-end networks are trained to learn these steps jointly, yielding better results \cite{shaked2017improved,mayer2016large,kendall2017end,ilg2017flownet,pang2017cascade,liang2017learning,gidaris2017detect}. 
However all these deep learning methods rely on a strong supervised component. As a consequence, they outperform traditional handcrafted optimization schemes only when a lot of ground-truth depth data is available, which is not the case in active stereo settings.

\textbf{Self-supervised passive stereo} is a possible solution for absence of ground-truth training data.   For instance when multiple images of the same scene are available, the images can warp between cameras using the estimated/calibrated pose and the depth, and the
loss between the reconstruction and the raw image can be used to train depth estimation systems without ground truth.
Taking advantage of spatial and temporal coherence, depth estimation algorithms can be trained unsupervised using monocular images \cite{godard2017unsupervised,garg2016unsupervised,kuznietsov2017semi}, video \cite{xie2016deep3d,zhou2017unsupervised}, and stereo \cite{zhong2017self}.
However, their results are blurry and far from comparable with supervised methods due to the required strong regularization such as left-right check \cite{godard2017unsupervised,zhong2017self}.  Also, they struggle in textureless and dark regions, as do all passive  methods.


\textbf{Active stereo} is an extension of the traditional passive stereo approach in which a texture is projected into the scene with an IR projector and cameras are augmented to perceive IR as well as visible spectra \cite{konolige2010projected}.  Intel R200 was the first attempt of commercialize an active stereo sensor, however its accuracy is poor compared to (older) structured light sensors, such as Kinect V1 \cite{hyperdepth,fanello2017ultrastereo}. Very recently, Intel released the D400 family \cite{inteld415,inteld435}, which provides higher resolution, $1280 \times 720$, and therefore has the potential to deliver more accurate depth maps.  The build-in stereo algorithm in these cameras uses a handcrafted binary descriptor (CENSUS) in combination with a semi-global matching scheme \cite{intelcams}.  It offers reasonable performance in a variety of settings \cite{sos}, but still suffers from common stereo matching issues addressed in this paper (edge fattening, quadratic error, occlusions, holes, etc.).

\textbf{Learning-based solutions for active stereo} are limited.  Past work has employed shallow architectures to learn a feature space where the matching can be performed efficiently \cite{fanello2017ultrastereo,fanello17_hashmatch,patchCollider}, trained a regressor to infer disparity \cite{hyperdepth}, or learned a direct mapping from pixel intensity to depth \cite{fanello14a}.   These methods fail in general scenes \cite{fanello14a}, suffer from interference and per-camera calibration \cite{hyperdepth}, and/or do not work well in texture-less areas due to their shallow descriptors and local optimization schemes \cite{fanello2017ultrastereo,fanello17_hashmatch}.   Our paper is the first to investigate how to design an end-to-end deep network for active stereo.

\begin{figure}[t]
    \centering
      \includegraphics[width=\columnwidth]{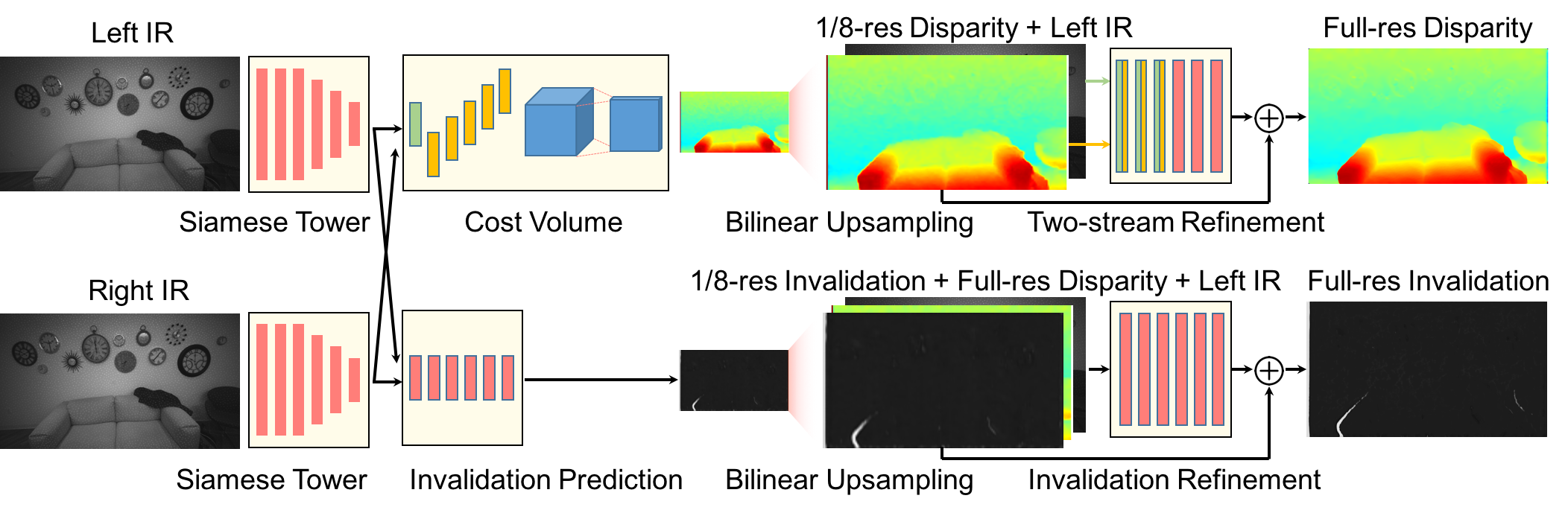}
    \caption{ActiveStereoNet architecture. We use a two stage network where a low resolution cost volume is built and infers the first disparity estimate. A bilinear upsampling followed by a residual network predicts the final disparity map. An ``Invalidation Network'' (bottom) is also trained end-to-end to predict a confidence map.}
 \label{fig:pipeline}
 \end{figure}

\section{Method}
In this section, we introduce the network architecture and training procedure for ActiveStereoNet.
The input to our algorithm is a rectified, synchronized pair of images with active illumination (see Fig. \ref{fig:teaser}), and the output is a pair of disparity maps at the original resolution.  For our experiments, we use the recently released Intel Realsense D435 that provides synchronized, rectified $1280 \times 720$ images at $30$fps. The focal length $f$ and the baseline $b$ between the two cameras are assumed to be known. Under this assumption, the depth estimation problem becomes a disparity search along the scan line. Given the output disparity $d$, the depth is obtained via $Z=\frac{bf}{d}$. 

Since no ground-truth training data is available for this problem, our main challenge is to train an end-to-end network that is robust to occlusion and illumination effects without direct supervision. The following details our algorithm.
 
\subsection{Network Architecture}
Nowadays, in many vision problems, the choice of the architecture plays a crucial role, and most of the efforts are spent in designing the right network. In active stereo, instead, we found that the most challenging part is the training procedure for a given deep network. In particular, since our setting is unsupervised, designing the optimal loss function has the highest impact on the overall accuracy. For this reason, we extend the network architecture proposed in \cite{stereonet}, which has shown superior performances in many passive stereo benchmarks. Moreover, the system is computationally efficient and allows us to run on full resolution at $60Hz$ on a high-end GPU, which is desirable for real-time applications.


The overall pipeline is shown in Fig. \ref{fig:pipeline}. We start from the high-resolution images and use a siamese tower to produces feature map in $1/8$ of the input resolution. We then build a low resolution cost volume of size $160 \times 90 \times 18$, allowing for a maximum disparity of $144$ in the original image, which corresponds to a minimum distance of $\sim 30$ cm on the chosen sensor.

The cost volume produces a downsampled disparity map using the soft argmin operator \cite{kendall2017end}. Differently from \cite{kendall2017end} and following \cite{stereonet} we avoid expensive 3D deconvolution and output a $160 \times 90$ disparity. This estimation is then upsampled using bi-linear interpolation to the original resolution ($1280 \times 720$). 
A final residual refinement retrieves the high-frequency details such as edges. 
Different from \cite{stereonet}, our refinement block starts with separate convolution layers running on the upsampled disparity and input image respectively, and merge the feature later to produce residual. This in practice works better to remove dot artifacts in the refined results.

Our network also simultaneously estimates an invalidation mask to remove uncertain areas in the result, which will be introduced in Sec. \ref{sec:invalidation}.


\subsection{Loss Function}
\label{sec:loss}
The architecture described is composed of a low resolution disparity and a final refinement step to retrieve high-frequency details. A natural choice is to have a loss function for each of these two steps. Unlike \cite{stereonet}, we are in an unsupervised setting due to the lack of ground truth data. A viable choice for the training loss $L$ then is the photometric error between the original pixels on the left image $I^{l}_{ij}$ and the reconstructed left image $\hat{I}^{l}_{ij}$, in particular $L = \sum_{ij} \Vert I^{l}_{ij} - \hat{I}^{l}_{ij} \Vert_1$. The reconstructed image $\hat{I}^l$ is obtained by sampling pixels from the right image $I^r$ using the predicted disparity $d$, i.e. $\hat{I}^{l}_{ij} =I^r_{i,j-d}$. Our sampler uses the Spatial Transformer Network (STN) \cite{stn}, which uses a bi-linear interpolation of 2 pixels on the same row and is fully differentiable. 

\begin{figure}[t]
    \centering
      \includegraphics[width=\columnwidth]{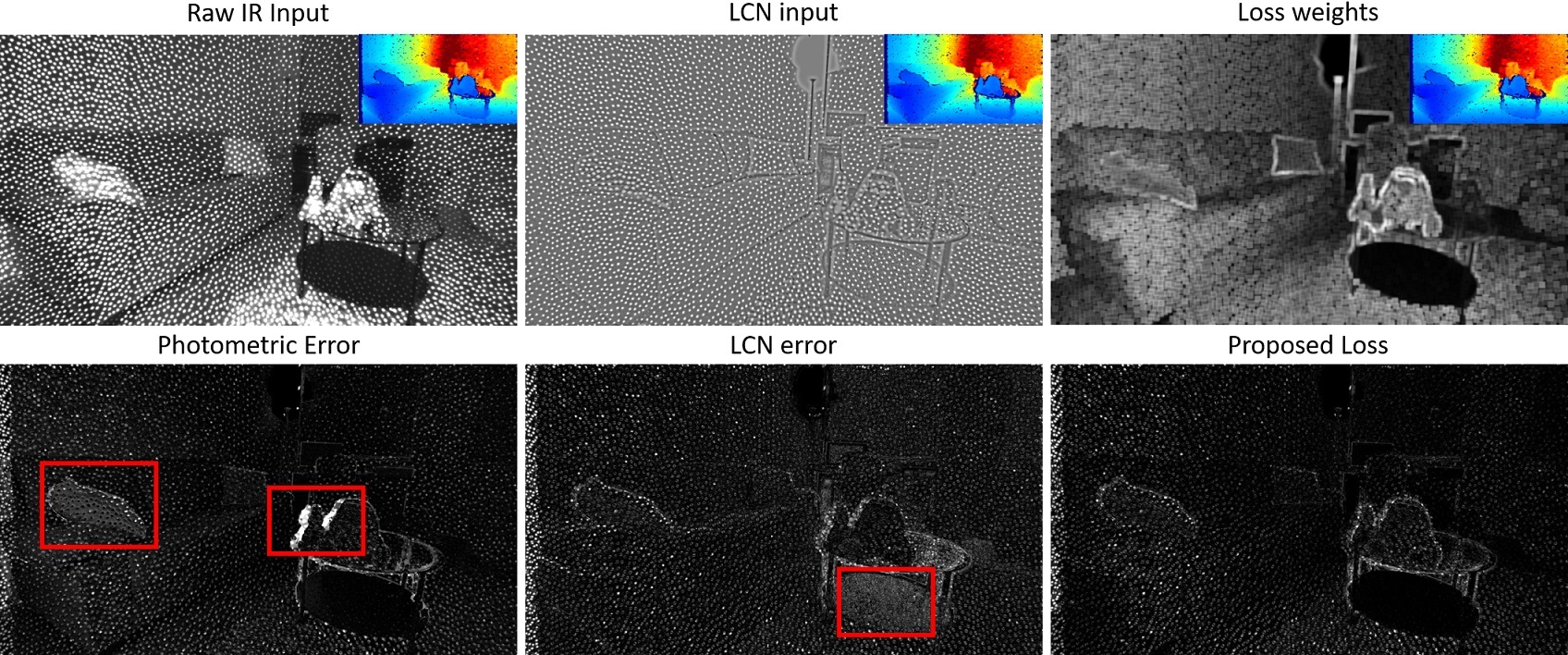}
    \caption{Comparisons between photometric loss (left), LCN loss (middle), and the proposed weighted LCN loss (right). Our loss is more robust to occlusions, it does not depend on the brightness of the pixels and does not suffer in low texture regions.}
 \label{fig:loss}
 \end{figure}
However, as shown in previous work \cite{loss_l2_reco}, the photometric loss is a poor choice for image reconstruction problems. This is even more dramatic when dealing with active setups. We recall that active sensors flood the scenes with texture and the intensity of the received signal follows the inverse square law $I \propto \frac{1}{Z^2}$, where $Z$ is the distance from the camera. In practice this creates an explicit dependency between the intensity and the distance (i.e. brighter pixels are closer). A second issue, that is also present in RGB images, is that the difference between two bright pixels is likely to have a bigger residual when compared to the difference between two dark pixels. Indeed if we consider image $I$, to have noise proportional to intensity \cite{noise_model}, the observed intensity for a given pixel can be written as: 
$I_{ij} = I^{\star}_{ij} + \mathcal{N}(0,\sigma_1I^{\star}_{ij} + \sigma_2),$
where $I^{\star}_{ij}$ is the noise free signal and the standard deviations $\sigma_1$ and $\sigma_2$ depend on the sensor \cite{noise_model}. It is easy to show that the difference between two correctly matched pixels $I$ and $\hat{I}$ has a residual:
$\epsilon = \mathcal{N}(0,\sqrt{(\sigma_1I^{\star}_{ij}+ \sigma_2)^2 +  (\sigma_3\hat{I}^{\star}_{ij}+ \sigma_4)^2} ),$
where its variance depends on the input intensities. This shows that for brighter pixels (i.e. close objects) the residual $\epsilon$ will be bigger compared to one of low reflectivity or farther objects. 

In the case of passive stereo, this could be a negligible effect, since in RGB images there is no correlation between intensity and disparity, however in the active case the aforementioned problem will bias the network towards closeup scenes, which will have always a bigger residual. The architecture will learn mostly those easy areas and smooth out the rest. The darker pixels, mostly in distant, requiring higher matching precision for accurate depth, however, are overlooked. In Fig. \ref{fig:loss} (left), we show the the reconstruction error for a given disparity map using the photometric loss. Notice how bright pixels on the pillow exhibits high reconstruction error due to the input dependent nature of the noise. 

An additional issue with this loss occurs in the occluded areas: indeed when the intensity difference between background and foreground is severe, this loss will have a strong contribution in the occluded regions, forcing the network to learn to fit those areas that, however, cannot really be explained in the data.

\paragraph{\textbf{Weighted Local Contrast Normalization.}} We propose to use a Local Contrast Normalization (LCN) scheme, that not only removes the dependency between intensity and disparity, but also gives a better residual in occluded regions. It is also invariant to brightness changes in the left and right input image. In particular, for each pixel, we compute the local mean $\mu$ and standard deviation $\sigma$ in a small $9\times9$ patch. These local statistics are used to normalize the current pixel intensity $I_{LCN} = \frac{I -\mu}{\sigma + \eta}$, where $\eta$ is a small constant. The result of this normalization is shown in Fig. \ref{fig:loss}, middle. Notice how the dependency between disparity and brightness is now removed, moreover the reconstruction error (Fig. \ref{fig:loss}, middle, second row) is not strongly biased towards high intensity areas or occluded regions. 

However, LCN suffers in low texture regions when the standard deviation $\sigma$ is close to zero (see the bottom of the table in Fig. \ref{fig:loss}, middle). Indeed these areas have a small $\sigma$ which will would amplify any residual together with noise between two matched pixels. To remove this effect, we re-weight the residual $\epsilon$ between two matched pixel $I_{ij}$ and $\hat{I}^{l}_{ij}$ using the local standard deviation $\sigma_{ij}$ estimated on the reference image in a $9 \times 9$ patch around the pixel $(i,j)$. In particular our reconstruction loss becomes:
$L = \sum_{ij}  \Vert \sigma_{ij} (I^{l}_{LCNij} - \hat{I}^{l}_{LCNij}) \Vert_1 = \sum_{ij} C_{ij}.$
Example of weights computed on the reference image are shown in Fig. \ref{fig:loss}, top right and the final loss is shown on the bottom right. Notice how these residuals are not biased in bright areas or low textured regions.

\subsection{Window-based Optimization}
\begin{figure}[t]
    \centering
      \includegraphics[width=\columnwidth]{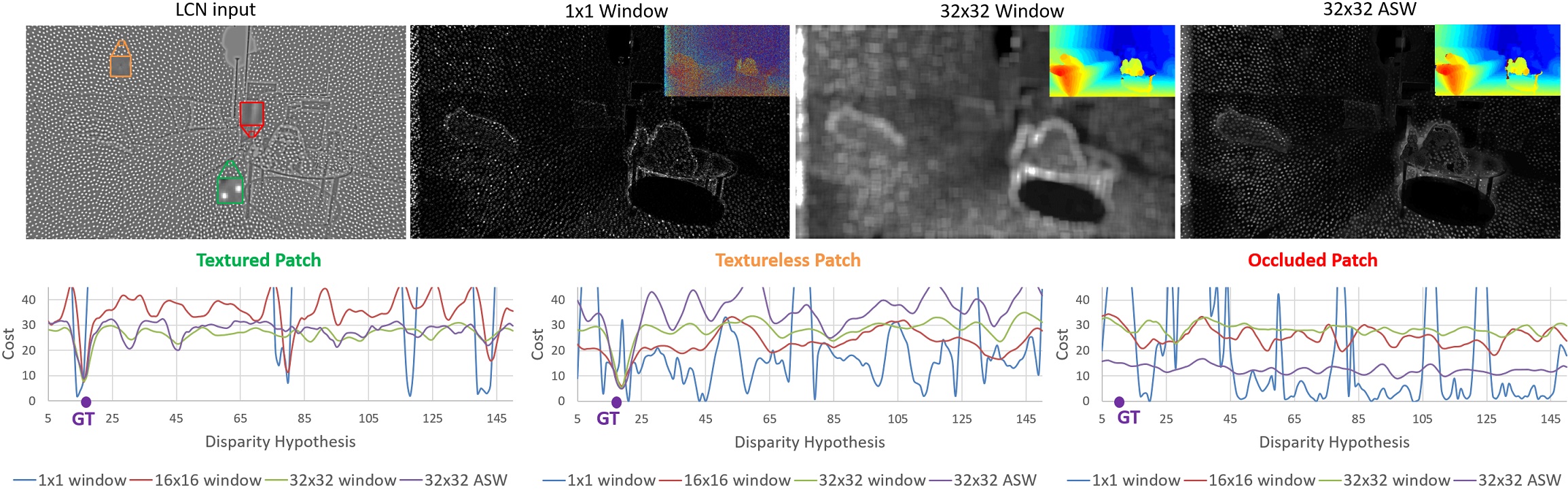}
    \caption{Cost volume analysis for a textured region (green), textureless patch (orange) and occluded pixel (red). Notice how the window size helps to resolve ambiguous (textureless) areas in the image, whereas in occluded pixels the lowest cost will always lead to the wrong solution. However large windows oversmooth the cost function and they do not preserve edges, where as the proposed Adaptive Support Weight loss aggregates costs preserving edges.}
 \label{fig:cost}
 \end{figure}
We now analyze in more details the behavior of the loss function for the whole search space. We consider a textured patch (green), a texture-less one (orange) and an occluded area (red) in an LCN image (see Fig. \ref{fig:cost}). We plot the loss function for every disparity candidate in the range of $[5,144]$. For a single pixel cost (blue curve), notice how the function exhibits a highly non-convex behavior (w.r.t. the disparity) that makes extremely hard to retrieve the ground truth value (shown as purple dots). Indeed a single pixel cost has many local minima, that could lie far from the actual optimum. In traditional stereo matching pipelines, a cost aggregation robustifies the final estimate using evidence from neighboring pixels. If we consider a window around each pixel and sum all the costs, we can see that the loss becomes smoother for both textured and texture-less patch and the optimum can be reached (see Fig. \ref{fig:cost}, bottom graphs). However as a drawback for large windows, small objects and details can be smooth out by the aggregation of multiple costs and cannot be recovered in the final disparity.

Traditional stereo matching pipelines aggregate the costs using an adaptive support (ASW) scheme \cite{asw}, which is very effective, but also slow hence not practical for real-time systems where approximated solutions are required \cite{asw_slow}. Here we propose to integrate the ASW scheme in the training procedure, therefore it does not affect the runtime cost. In particular, we consider a pixel $(i,j)$ with intensity $I_{ij}$ and instead of compute a per-pixel loss, we aggregate the costs $C_{ij}$ around a $2k \times 2k$ window following:
$\hat{C_{ij}} = \frac{\sum_{x=i-k}^{i+k-1} \sum_{y=j-k}^{j+k-1} w_{x,y} C_{ij}}{\sum_{x=i-k}^{i+k-1} \sum_{y=j-k}^{j+k-1} w_{x,y}},$
where $w_{xy} = \exp(-\frac{\vert I_{ij}- I_{xy}\vert}{\sigma_w})$, with $\sigma_w=2$. As shown in Fig. \ref{fig:cost} right, this aggregates the costs (i.e. it smooths the cost function), but it still preserves the edges. In our implementation we use a $32 \times 32$ during the whole training phase. We also tested a graduated optimization approach \cite{graduated1,graduated}, where we first optimized our network using $64 \times 64$ window
and then reduce it every $15000$ iterations by a factor of $2$, until we reach a single pixel loss. However this solution led to very similar results compared to a single pixel loss during the whole training.

\subsection{Invalidation Network}
\label{sec:invalidation}
So far the proposed loss does not deal with occluded regions and wrong matches (i.e. textureless areas). An occluded pixel does not have any useful information in the cost volume even when brute-force search is performed at different scales (see in Fig. \ref{fig:cost}, bottom right graph). To deal with occlusions, traditional stereo matching methods use a so called left-right consistency check, where a disparity is first computed from the left view point ($d_l$), then from the right camera ($d_r$) and invalidate those pixels with $\vert d_l - d_r \vert > \theta$. Related work use a left-right consistency in the loss minimization \cite{godard2017unsupervised}, however this leads to oversmooth edges which become flying pixels (outliers) in the pointcloud. Instead, we propose to use the left-check as a hard constraint by defining a mask for a pixel $(i,j)$: $m_{ij} = \vert d_l - d_r \vert < \theta$, with $\theta=1$ disparity. Those pixels with $m_{ij}=0$ are ignored in the loss computation. To avoid a trivial solution (i.e. all the pixels are invalidated), similarly to \cite{zhou2017unsupervised}, we enforce a regularization on the number of valid pixels by minimizing the cross-entropy loss with constant label $1$ in each pixel location. We use this mask in both the low-resolution disparity as well as the final refined one.

At the same time, we train an invalidation network (fully convolutional), that takes as input the features computed from the Siamese tower and produces first a low resolution invalidation mask, which is then upsampled and refined with a similar architecture used for the disparity refinement. This allows, at runtime, to avoid predicting the disparity from both the left and the right viewpoint to perform the left-right consistency, making the inference significantly faster. 

\section{Experiments}
\label{sec:experiments}
We performed a series of experiments to evaluate ActiveStereoNet (ASN).  In addition to analyzing the accuracy of depth predictions in comparison to previous work, we also provide results of ablation studies to investigate how each component of the proposed loss affects the results. In the supplementary material we also evaluate the applicability of our proposed self-supervised loss in passive (RGB) stereo, showing improved generalization capabilities and compelling results on many benchmarks.


\subsection{Dataset and Training Schema}
We train and evaluate our method on both real and synthetic data.

For the \textit{real dataset}, we used an Intel Realsense D435 camera \cite{inteld435} to collect $10000$ images for training in an office environment, plus $100$ images in other \textit{unseen} scenes for testing (depicting people, furnished rooms and objects).

For the \textit{synthetic dataset}, we used Blender to render IR and depth images of indoor scenes such as living rooms, kitchens, and bedrooms, as in \cite{fanello2017ultrastereo}.  Specifically, we render synthetic stereo pairs with $9$ cm baseline using projective textures to simulate projection of the Kinect V1 dot pattern onto the scene. We randomly move the camera in the rendered rooms and capture left IR image, right IR image as well as ground truth depth. Examples of the rendered scenes are showed in Fig. \ref{fig:real_synthetics}, left. The synthetic training data consists of $10000$ images and the test set is composed of $1200$ frames comprehending new scenes.

For both real and synthetic experiments, we trained the network using RMSprop \cite{rmsprop}. We set the learning rate to $1\mathrm{e}{-4}$ and reduce it by half at $\frac{3}{5}$ iterations and to a quarter at $\frac{4}{5}$ iterations. We stop the training after $100000$ iterations, that are usually enough to reach the convergence. Although our algorithm is self-supervised, we \textit{did not} fine-tune the model on any of the test data since it reduces the generalization capability in real applications.

\subsection{Stereo Matching Evaluation}
In this section, we compare our method on real data with state of the art stereo algorithms qualitatively and quantitatively using traditional stereo matching metrics, such as jitter and bias.

\paragraph{\textbf{Bias and Jitter.}} It is known that a stereo system with baseline $b$, focal length $f$, and a subpixel disparity precision of $\delta$, has a depth error $\epsilon$ that increases quadratically with respect to the depth $Z$ according to $\epsilon=\frac{\delta Z^2}{bf}$ \cite{Szeliski_book}. 
Due to the variable impact of disparity error on the depth, naive evaluation metrics, like mean error of disparity, does not effectively reflect the quality of the estimated depth. In contrast, we first show error of depth estimation and calculate corresponding error in disparity.

 \begin{figure}[t]
    \centering
      \includegraphics[width=0.9\columnwidth]{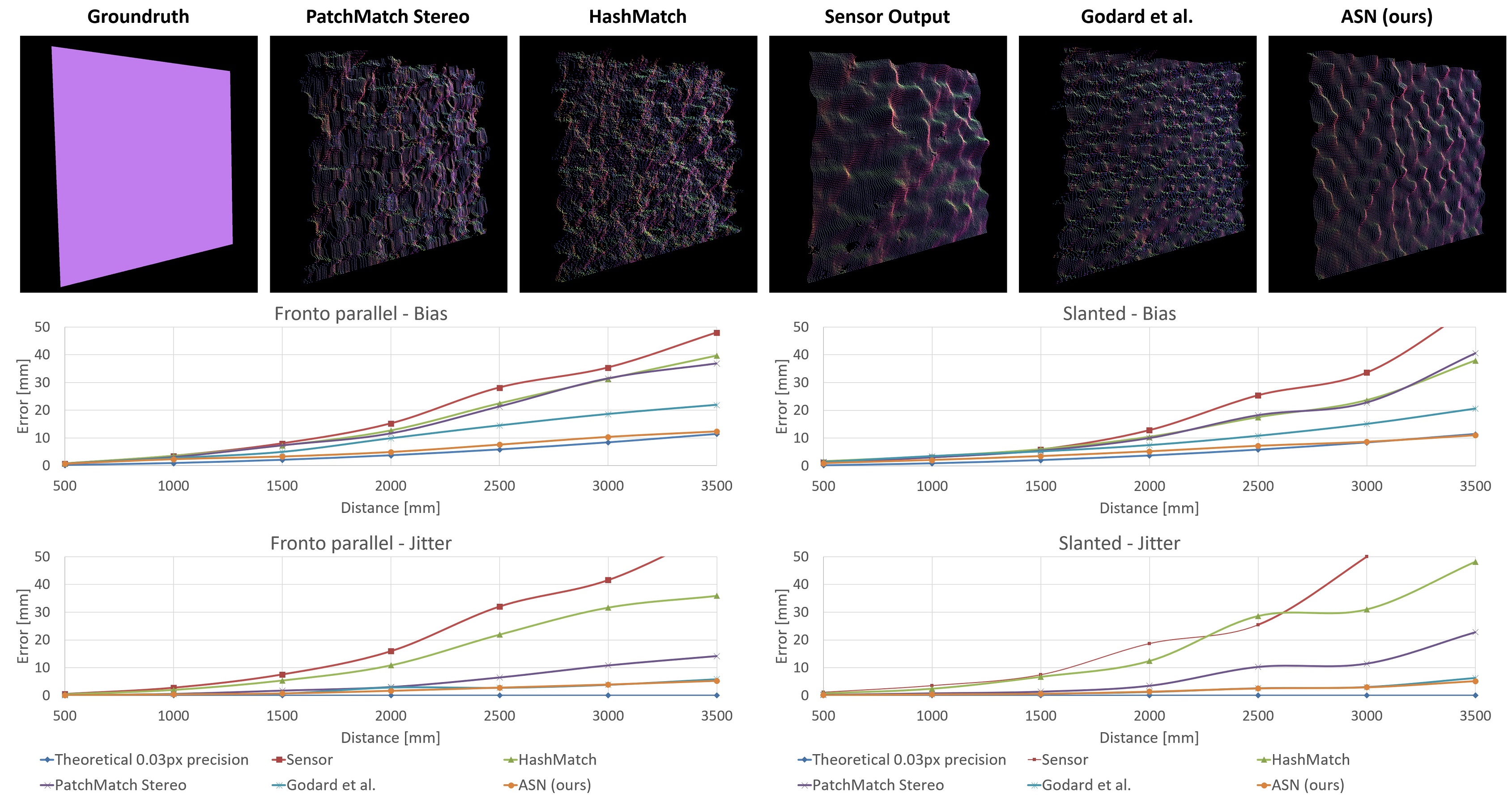}
    \caption{Quantitative Evaluation with state of the art. We achieve one order of magnitude less bias with a subpixel precision of $0.03$ pixels with a very low jitter (see text). We also show the predicted pointclouds for various methods of a wall at $3000$mm distance. Notice that despite the large distance ($3$m), our results is the less noisy compared to the considered approaches.}
 \label{fig:quantitative_active}
 \end{figure}
 
To assess the subpixel precision of ASN, we recorded 100 frames with the camera in front of a flat wall at distances ranging from $500$ mm to $3500$ mm, and also 100 frames with the camera facing the wall at an angle of $50 \deg$ to assess the behavior on slanted surfaces.  In this case, we evaluate by comparing to ``ground truth'' obtained with robust plane fitting.  

To characterize the precision, we compute \textit{bias} as the average $\ell_1$ error between the predicted depth and the ground truth plane.
Fig. \ref{fig:quantitative_active} shows the bias with regard to the depth for our method, sensor output \cite{intelcams}, the state of the art local stereo methods (PatchMatch \cite{bleyer2011patchmatch}, HashMatch \cite{fanello17_hashmatch}), and our model trained using the state of the art unsupervised loss \cite{godard2017unsupervised}, together with visualizations of point clouds colored by surface normal.
Our system performs significantly better than the other methods at all distances, and its error does not increase dramatically with depth.
The corresponding subpixel disparity precision of our system is $1/30th$ of a pixel, which is obtained by fitting a curve using the above mentioned equation (also shown in Fig. \ref{fig:quantitative_active}).
This is one order of magnitude lower than the other methods where the precision is not higher than $0.2$ pixel.

To characterize the noise, we compute the \textit{jitter} as the standard deviation of the depth error.  Fig. \ref{fig:quantitative_active} shows that our method achieves the lowest jitter at almost every depth in comparison to other methods.

\begin{figure}[t]
    \centering
      \includegraphics[width=\columnwidth]{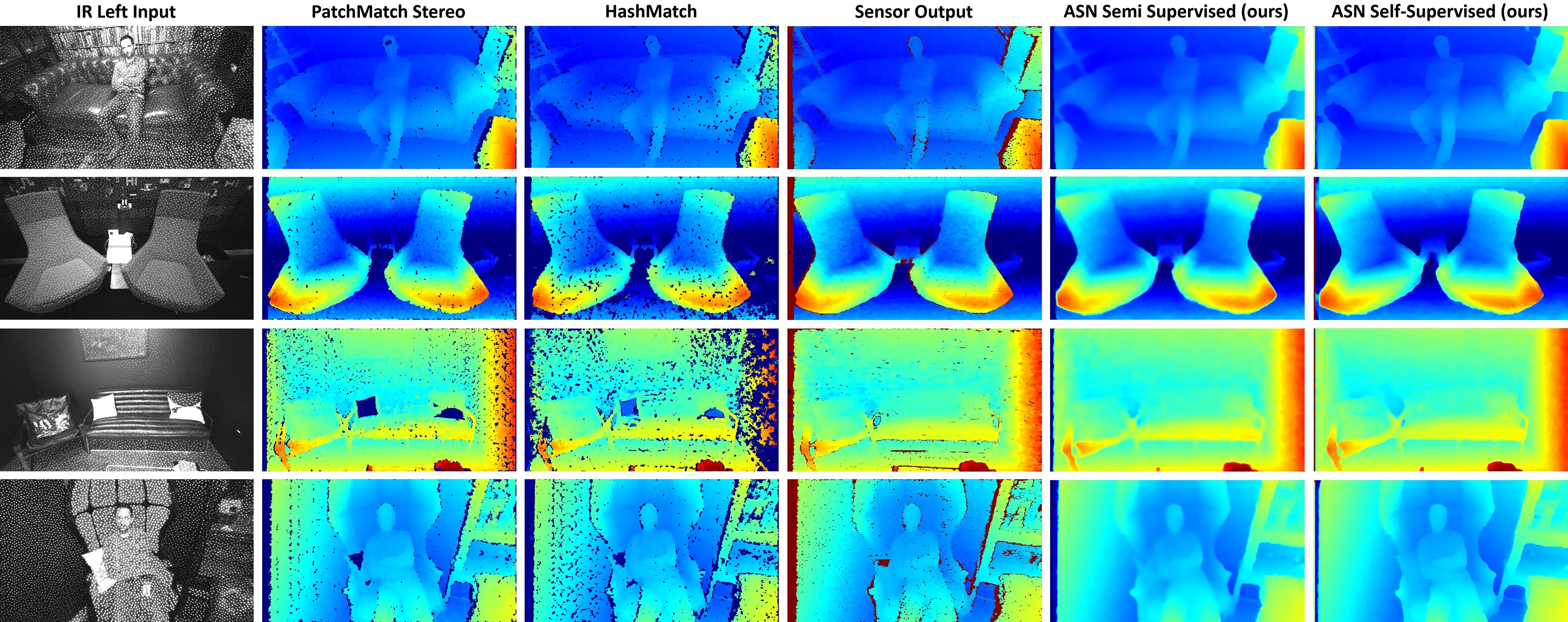}
    \caption{Qualitative Evaluation with state of the art. Our method produces detailed disparity maps. State of the art local methods \cite{bleyer2011patchmatch,fanello17_hashmatch} suffer from textureless regions. The semi-global scheme used by the sensor \cite{intelcams} is noisier and it oversmooths the output.}
 \label{fig:qualitative_active}
 \end{figure}
\paragraph{\textbf{Comparisons with State of the Art.}}
More qualitative evaluations of ASN in challenging scenes are shown in Fig. \ref{fig:qualitative_active}. As can be seen, local methods like PatchMatch stereo \cite{bleyer2011patchmatch} and HashMatch \cite{fanello17_hashmatch} do not handle mixed illumination with both active and passive light, and thus produce incomplete disparity images (missing pixels shown in black).
The sensor output using a semi-global scheme is more suitable for this data \cite{intelcams}, but it is still susceptible to image noise (note the noisy results in the fourth column).
In contrast, our method produces complete disparity maps and preserves sharp boundaries. 

More examples on real sequences are shown in Fig. \ref{fig:real_synthetics} (right), where we show point clouds colored by surface normal. Our output preserves all the details and exhibits a low level of noise. In comparison, our network trained with the self-supervised method by Godard et al. \cite{godard2017unsupervised} over-smooths the output, hallucinating geometry and flying pixels. Our results are also free from the texture copying problem, most likely because we use a cost volume to explicitly model the matching function rather than learn directly from pixel intensity. Even though the training data is mostly captured from office environment, we find ASN generalize well to various testing scenes, e.g. living room, play room, dinning room, and objects, e.g. person, sofas, plants, table, as shown in figures.

\subsection{Ablation Study}
In this section, we evaluate the importance of each component in the ASN system. Due to the lack of ground truth data, most of the results are qualitative -- when looking at the disparity maps, please pay particular attention to noise, bias, edge fattening, flying pixels, resolution, holes, and generalization capabilities. 

\begin{figure}[t]
    \centering
      \includegraphics[width=\columnwidth]{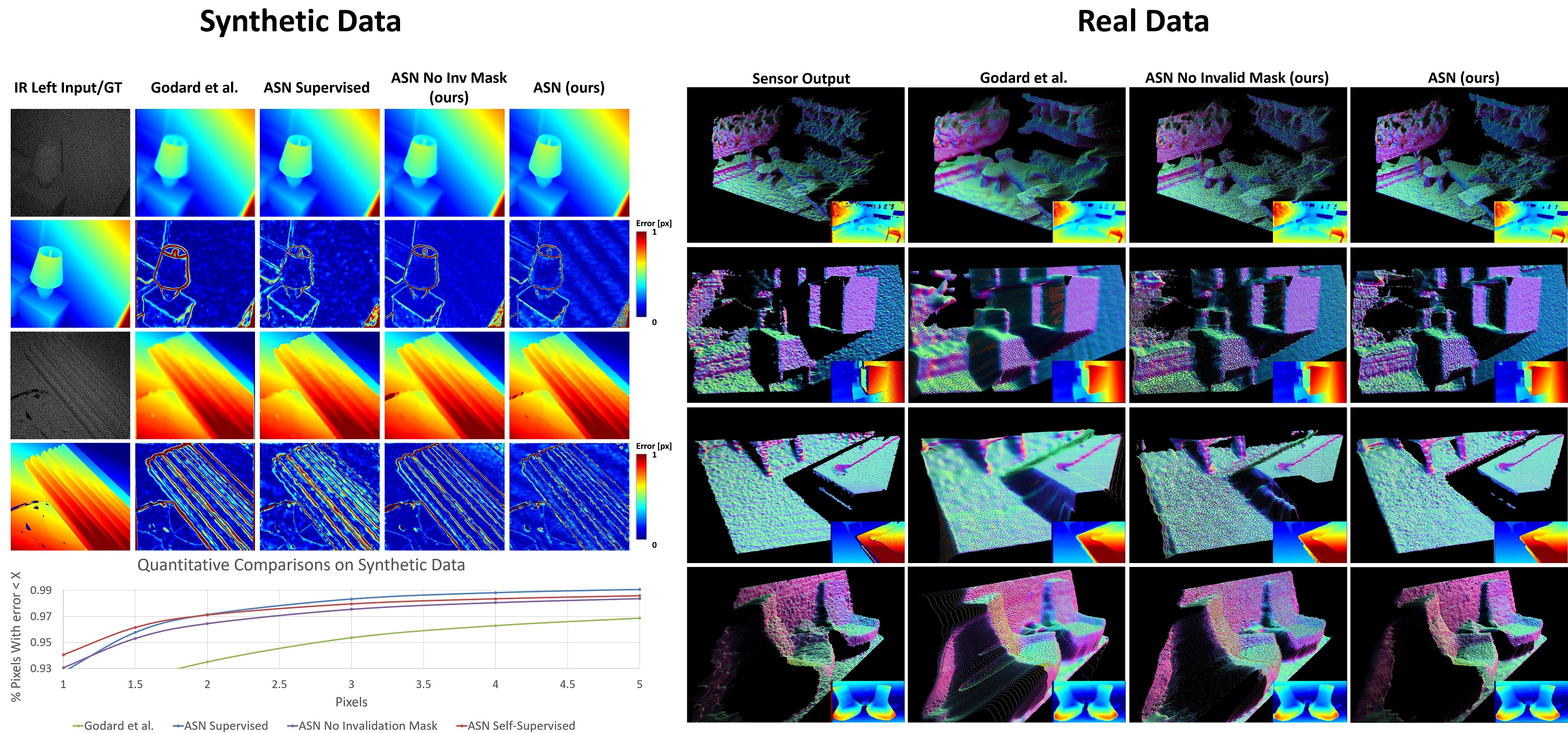}
    \caption{Evaluation on Synthetic and Real Data. On synthetic data (left), notice how our method has the highest percentage of pixels with error smaller than $1$ disparity. We also produce sharper edges and less noisy output compared to other baselines. The state of the art self-supervised method by Godard et al. \cite{godard2017unsupervised} is very inaccurate near discontinuities. On the right, we show real sequences from an Intel RealSense D435 where the gap between \cite{godard2017unsupervised} and our method is even more evident: notice flying pixels and oversmooth depthmaps produced by Godard et al. \cite{godard2017unsupervised}. Our results has higher precision than the sensor output.}
 \label{fig:real_synthetics}
 \end{figure}

\paragraph{\textbf{Self-supervised vs Supervised.}}
Here we perform more evaluations of our self-supervised model on synthetic data when supervision is available as well as on real data using the depth from the sensor as supervision (together with the proposed loss). Quantitative evaluation on synthetic data (Fig. \ref{fig:real_synthetics}, left bottom), shows that the supervised model (blue) achieves a higher percentage of pixels with error less than $5$ disparity, however for more strict requirements (error less than 2 pixels) our self-supervised loss (red) does a better job.
This may indicate overfitting of the supervised model on the training set. This behavior is even more evident on real data: the model was able to fit the training set with high precision, however on test images it produces blur results compared to the self-supervised model (see Fig. \ref{fig:qualitative_active}, ASN Semi Supervised vs ASN Self-Supervised). 

\begin{figure}[t]
    \centering
      \includegraphics[width=\columnwidth]{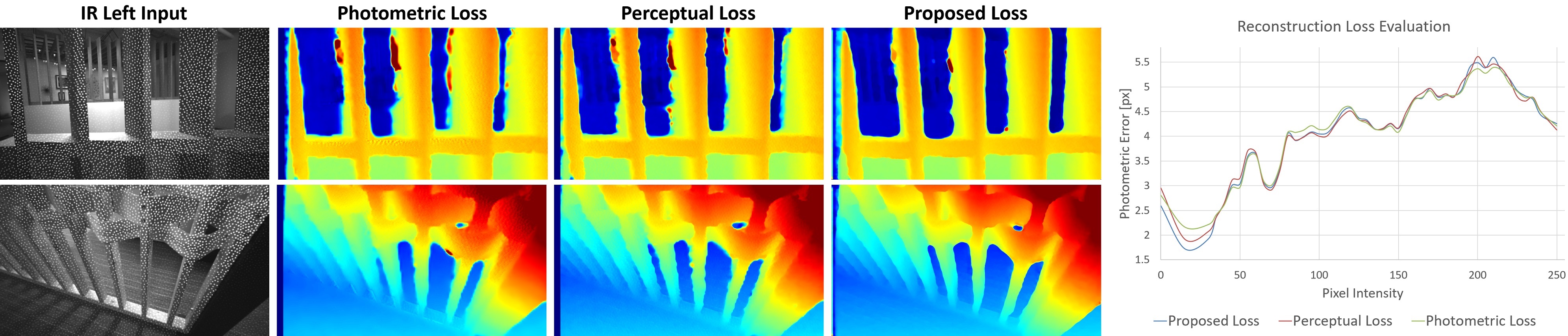}
    \caption{Ablation study on reconstruction loss. Same networks, trained on $3$ different reconstruction losses. Notice how the proposed WLCN loss infers disparities that better follow the edges in these challenging scenes. Photometric and Perceptual losses have also a higher level of noise. On the right, we show how our loss achieves the lowest reconstruction error for low intensity pixels thanks to the proposed WLCN.}
 \label{fig:qualitative_photo_lcn_perc}
 \end{figure}
\paragraph{\textbf{Reconstruction Loss.}}
We next investigate the impact of our proposed WLCN loss (as described in Sec. \ref{sec:loss}) in comparison to a standard photometric error (L1) and a perceptual loss \cite{perceptual} computed using feature maps from a pre-trained VGG network. In this experiment, we trained three networks with the same parameters, changing only the reconstruction loss: photometric on raw IR, VGG conv-1, and the proposed WLCN, and investigate their impacts on the results.

\begin{figure}[t]
    \centering
      \includegraphics[width=\columnwidth]{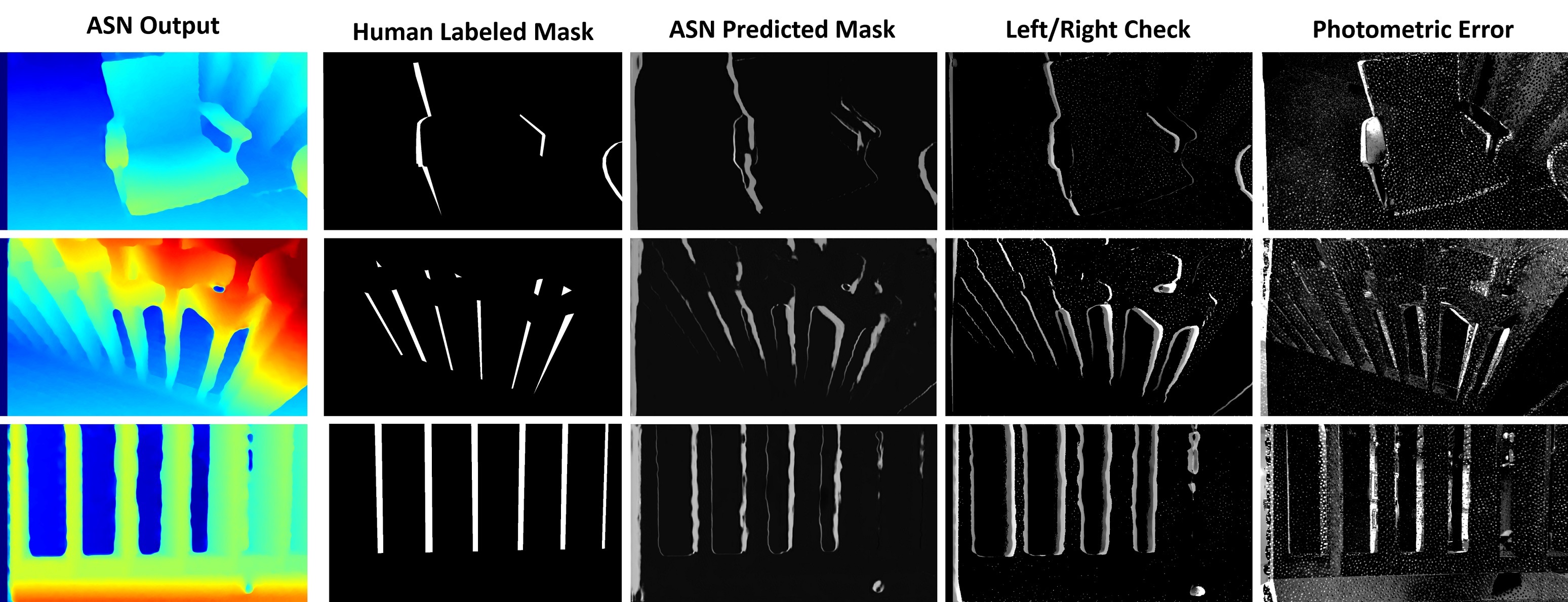}
    \caption{Invalidation Mask prediction. Our invalidation mask is able to detect occluded regions and it reaches an average precision of $80.7 \%$ (see text).}
 \label{fig:occlusion}
 \end{figure}

To compute accurate metrics, we labeled the occluded regions in a subset of our test case manually (see Fig. \ref{fig:occlusion}). For those pixels that were not occluded, we computed the photometric error of the raw IR images given the predicted disparity image. In total we evaluated over $10$M pixels.
In Fig. \ref{fig:qualitative_photo_lcn_perc} (right), we show the photometric error of the raw IR images for the three losses with respect to the pixel intensities. The proposed WLCN achieves the lowest error for small intensities, showing that the loss is not biased towards bright areas. For the rest of the range the losses get similar numbers. Please notice that our loss achieves the lowest error even we did not explicitly train to minimize the photometric reconstruction. Although the numbers may seem similar, the effect on the final disparity map is actually very evident. We show some examples of predicted disparities for each of the three different losses in Fig. \ref{fig:qualitative_photo_lcn_perc} (left). Notice how the proposed WLCN loss suffers from less noise, produces crisper edges, and has a lower percentage of outliers.  In contrast, the perceptual loss highlights the high frequency texture in the disparity maps (i.e. dots), leading to noisy estimates. Since VGG conv-1 is pre-trained, we observed that the responses are high on bright dots, biasing the reconstruction error again towards close up scenes. We also tried a variant of the perceptual loss by using the output from our Siamese tower as the perceptual feature, however the behavior was similar to the case of using the VGG features.

\paragraph{\textbf{Invalidation Network.}}
We next investigate whether excluding occluded region from the reconstruction loss is important to train a network -- i.e., to achieve crisper edges and less noisy disparity maps.  We hypothesize that the architecture would try to overfit occluded regions  without this feature (where there are no matches), leading to higher errors throughout the images.  We test this quantitatively on synthetic images by computing the percentage of pixels with disparity error less than $x \in [1,5]$.   The results are reported in Fig. \ref{fig:real_synthetics}.
With the invalidation mask employed, our model outperforms the case without for all the error threshold (Red v.s Purple curve, higher is better).
We further analyze the produced disparity and depth maps on both synthetic and real data. On synthetic data, the model without invalidation mask shows gross error near the occlusion boundary (Fig. \ref{fig:real_synthetics}, left top).
Same situation happens on real data (Fig. \ref{fig:real_synthetics}, right), where more flying pixels exhibiting when no invalidation mask is enabled. 

As a byproduct of the invalidation network, we obtain a confidence map for the depth estimates. In Fig. \ref{fig:occlusion} we show our predicted masks compared with the ones predicted with a left-right check and the photometric error. To assess the performances, we used again the images we manually labeled with occluded regions and computed the average precision (AP). Our invalidation network and left right check achieved the highest scores with an AP of $80.7 \%$ and $80.9\%$ respectively, whereas the photometric error only reached $51.3 \%$. We believe that these confidence maps could be useful for many higher-level applications.

\begin{figure}[t]
    \centering
      \includegraphics[width=\columnwidth]{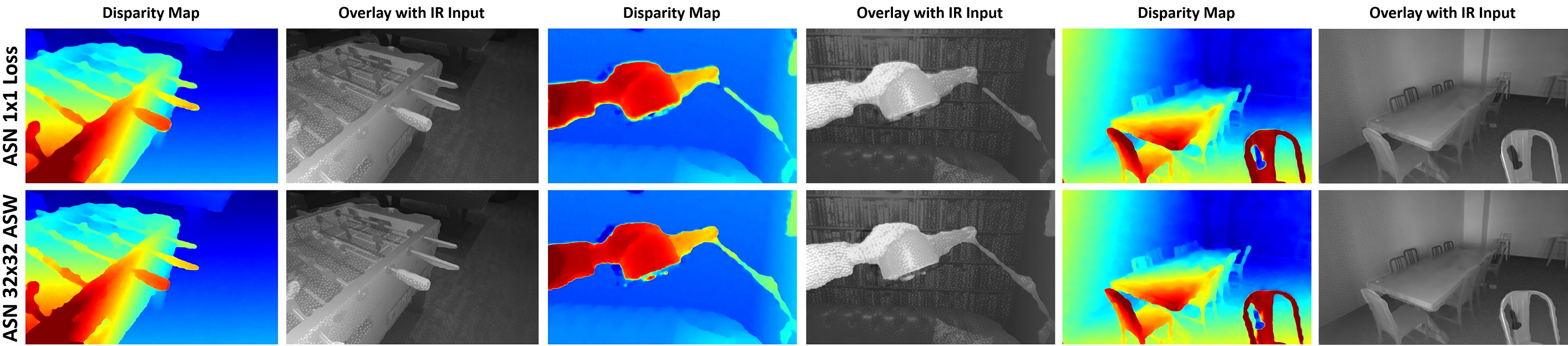}
    \caption{Comparison between single pixel loss and the proposed window based optimization with adaptive support scheme. Notice how the ASW is able to recover more thin structures and produce less edge fattening.}
 \label{fig:window_based_optimization}
 \end{figure}
\paragraph{\textbf{Window based Optimization.}}
The proposed window based optimization with Adaptive Support Weights (ASW) is very important to get more support for thin structures that otherwise would get a lower contribution in the loss and treated as outliers. We show a comparison of this in Fig. \ref{fig:window_based_optimization}. Notice how the loss with ASW is able to recover hard thin structures with higher precision. Moreover, our window based optimization also produces smoother results while preserving edges and details. Finally, despite we use a window-based loss, the proposed ASW strategy has a reduced amount of edge fattening.

\section{Discussion}
We presented ActiveStereoNet (ASN) the first deep learning method for active stereo systems. We designed a novel loss function to cope with high-frequency patterns, illumination effects, and occluded pixels to address issues of active stereo in a self-supervised setting. We showed that our method delivers very precise reconstructions with a subpixel precision of $0.03$ pixels, which is one order of magnitude better than other active stereo matching methods. Compared to other approaches, ASN does not oversmooth details, and it generates complete depthmaps, crisp edges, and no flying pixels. As a byproduct, the invalidation network is able to infer a confidence map of the disparity that can be used for high level applications requiring occlusions handling. Numerous experiments show state of the art results on different challenging scenes with a runtime cost of $15$ms per frame using an NVidia Titan X. 

\paragraph{\textbf{Limitations and Future Work.}}
Although our method generates compelling results there are still issues with transparent objects and thin structures due to the low resolution of the cost volume. In future work, we will propose solutions to handle these cases with high level cues, such as semantic segmentation. 

\clearpage

\bibliographystyle{splncs}
\bibliography{egbib}

\clearpage
\appendix

\title{ActiveStereoNet: End-to-End Self-Supervised Learning for Active Stereo Systems\\(Supplementary Materials)} 
\author{Yinda Zhang$^{1,2}$, Sameh Khamis$^1$, Christoph Rhemann$^1$, Julien Valentin$^1$, Adarsh Kowdle$^1$, Vladimir Tankovich$^1$, Michael Schoenberg$^1$, \\ Shahram Izadi$^1$, Thomas Funkhouser$^{1,2}$, Sean Fanello$^1$   }
\institute{$^1$Google Inc., $^2$Princeton University 
}
\maketitle
In this supplementary material, we provide additional information regarding our implementation details and more ablation studies on important components of the proposed framework. We conclude the supplementary material evaluating our method on passive stereo matching, showing that the proposed approach can be applied also on RGB images.

\section{Additional Implementation Details}
To ensure reproducibility, in the following, we give all the details needed to re-implement the proposed architecture.

\begin{figure}[t]
    \centering
      \includegraphics[width=\columnwidth]{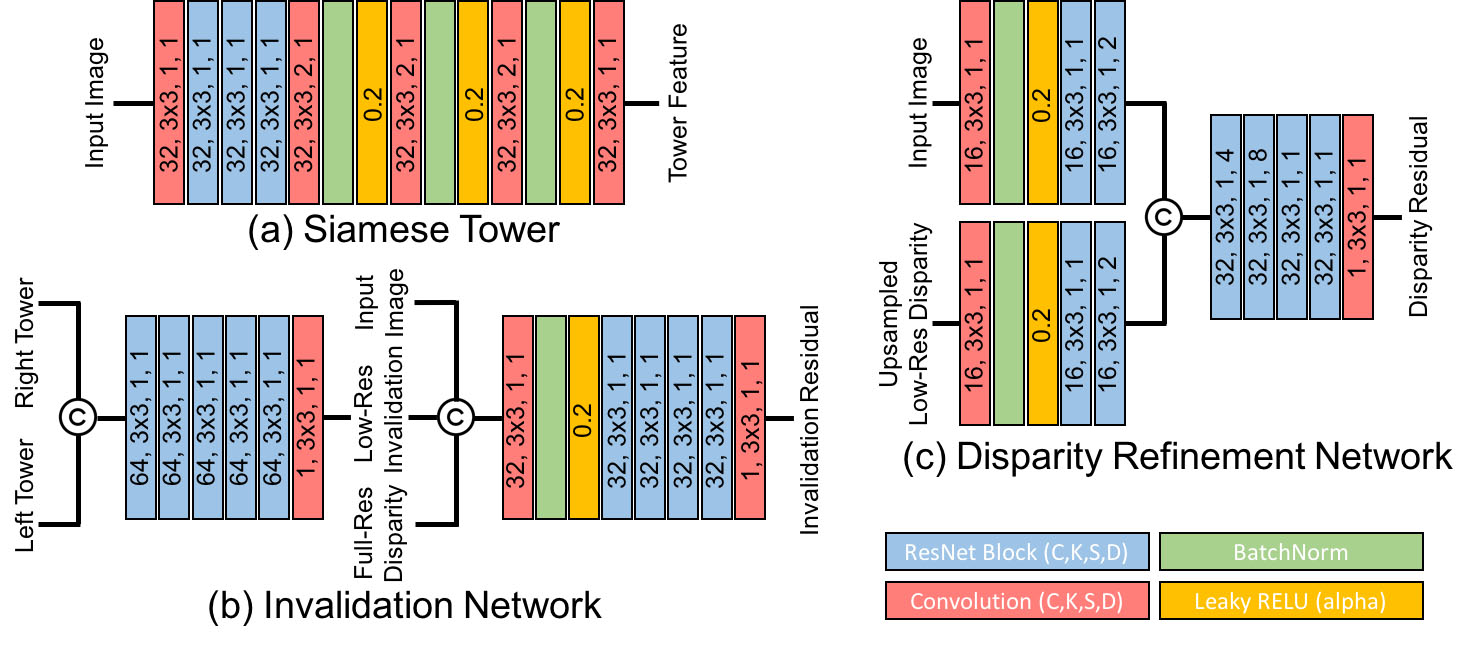}
    \caption{Detailed Network Architecture. (a) Siamese Tower, (b) Invalidation Network, (c) Disparity Refinement Network. In resnet block and convolution, the numbers are number of channels, kernel size, stride, and dilated rate. In leaky RELU, the number is the slope for $x<0$. $\copyright$ means feature map concatenation.}
 \label{fig:detail}
 \end{figure}

\subsection{Architecture}
\label{sec:architecture2}
Our model extends the architecture of \cite{stereonet}. Although the specific architecture is out of the scope of this paper, we introduced three different modifications to address specific problems of active stereo matching.

First, we adjust Siamese tower to maintain more high frequency signals from the input image. Unlike \cite{stereonet} that aggressively reduce resolution at the beginning of the tower, we run several residual blocks on full resolution and reduce resolution later using convolution with stride (Fig. \ref{fig:detail} (a)).

Second, we use a two-stream refinement network (Fig. \ref{fig:detail} (c)). Instead of stacking the image and the upsampled low resolution disparity, we feed these two into different pathways, where each pathway consists of 1 layer of convolution with 16 channels and 3 resnet blocks that maintain the number channels. The output of two pathways are concatenated together into a 32-dim feature map, which is further fed into 3 resnet blocks. We found this architecture produces results with way less dot artifacts.

Lastly, we have an additional invalidation network which predicts the confidence of the estimated disparity (Fig. \ref{fig:detail} (b)). One pursuing only very accurate depth could use this output to remove unwanted depth.
The invalidation network takes the concatenation of the left and right tower features, and feed them into 5 resnet and 1 convolution in the end. The output is an invalidation map with the same resolution of the tower feature, i.e. $1/8$ of the input resolution.
This low resolution invalidation map is upsampled to full resolution by bilinear interpolation.
In order to refine it, we concatenate this upsampled invalidation map with the full resolution refined disparity and the input image, and feed them into 1 conv + 4 resnet block + 1 conv.
The output is a one dimensional score map with higher value representing invalidate area.

During the training, we formulate the invalidation mask learning as a classification task. Note that the problem is completely unbalanced, i.e. most of the pixels are valid, therefore a trivial solution for the network consists of assigning all the pixels a positive label. To avoid this, we reweigh the output space by a factor of $10$ (we notice that invalid pixels are roughly $10 \%$ of the valid ones). For valid pixels we assign the value $y^+=1$, whereas for invalid pixels we use $y^-=-10$. The final loss is defined as an $\ell$-1 loss between the estimation and the ground truth mask\footnote{We also tried other classification losses, like the logistic loss, and they all led to very similar results.}.

\subsection{Training}
\subsubsection{Resolution}
Train/test on full resolution is important for active stereo system to get accurate depth (Sec. \ref{sec:resolution}). However, during the training, the model cannot fit in a 12GB memory for a full resolution of $1280\times720$.
To still enable the training in full resolution, we randomly pick a region with $1024\times 256$ pixels, and crop from this same area from left and right images.
This does not change the disparity, and thus models trained on small disparity can directly work on full resolution during the test.
\subsubsection{Invalidation Network}
Our invalidation network is trained fully self-supervised. The ``ground-truth'' is generated on the fly by using a hard left-right consistency check with a disparity threshold equal to $1$. However, we found this function to be very unstable at the beginning of the training, since the network has not converged yet. 
To regularize its behavior, we first disable the invalidation network and only train the disparity network for $20000$ iterations. After that we enable also the invalidation network. In practice we found this strategy is helpful for the disparity network to converge and prevent invalidation network from affecting the disparity.
\begin{figure}[t]
    \centering
      \includegraphics[width=\columnwidth]{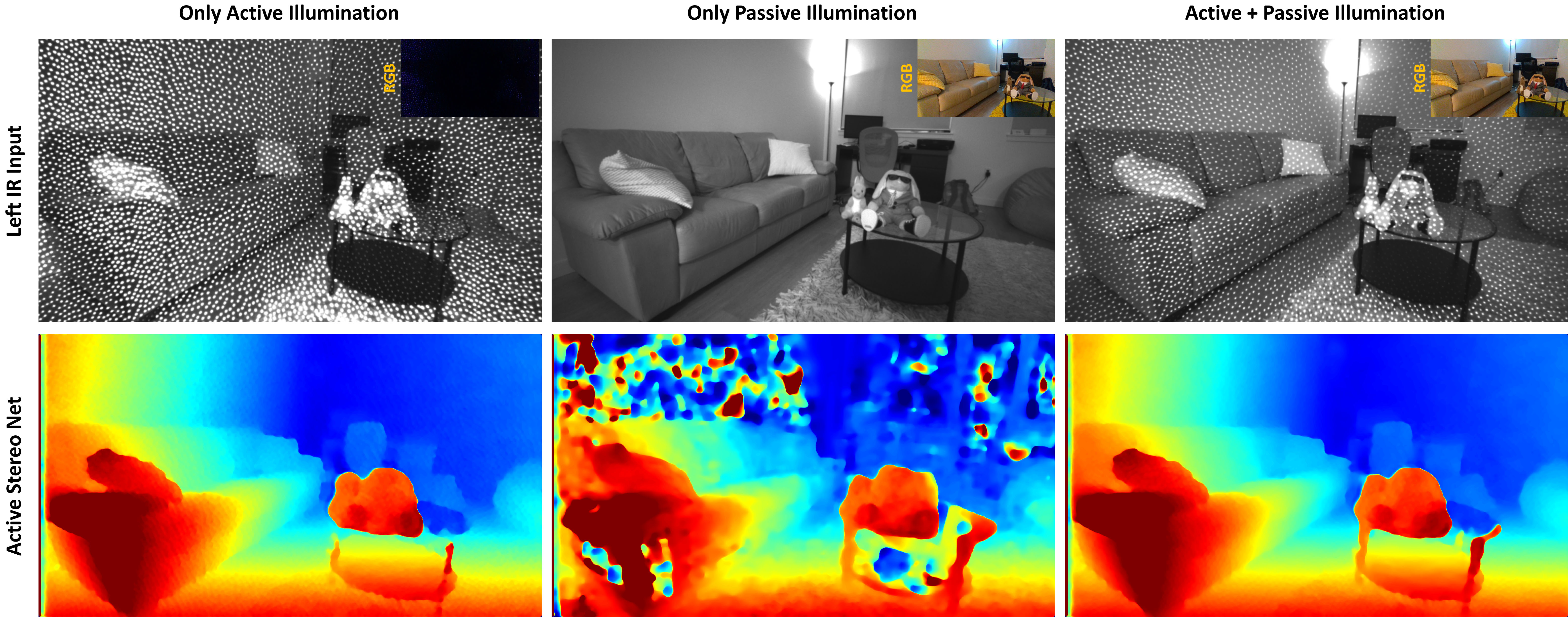}
    \caption{Active vs Passive Illumination. Notice the importance of both: only active illumination exhibits higher level or noise,  passive illumination struggles in textureless regions. When both the illumination are present we predict high quality disparity maps.}
 \label{fig:actrive_passive}
 \end{figure}
 \begin{figure}[t]
    \centering
      \includegraphics[width=\columnwidth]{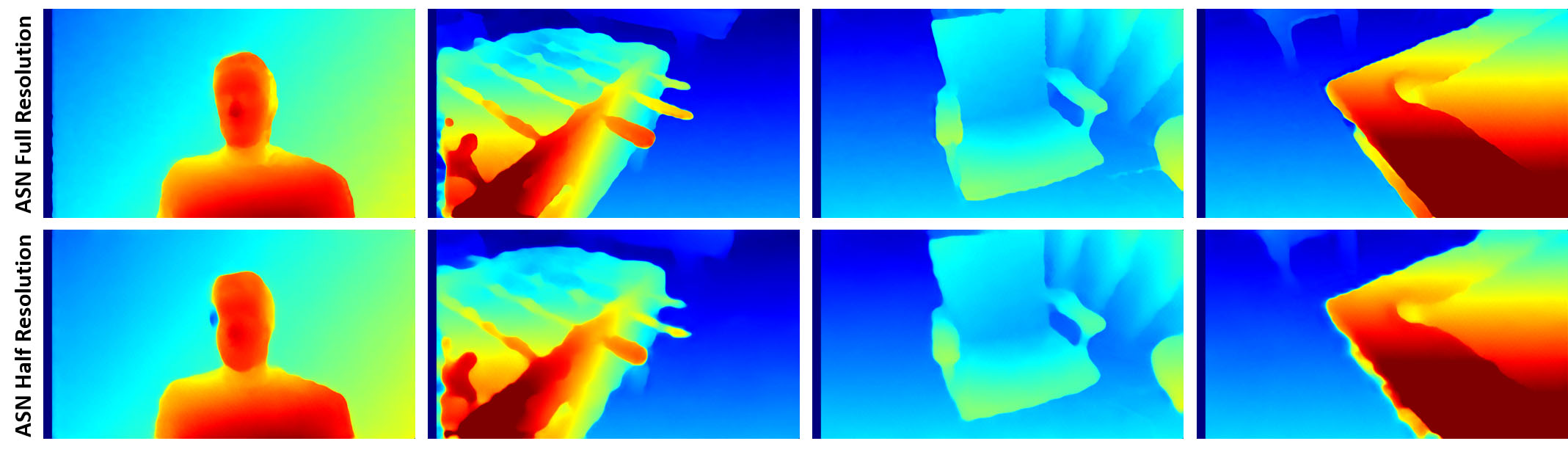}
    \caption{Qualitative Evaluation Resolution. Notice how full resolution is needed to produce thin structures and crisp edges.}
 \label{fig:qual_resolution}
 \end{figure}
 
\section{Additional Evaluations}
In this section we perform more ablation studies on different components of the network, we then show additional qualitative results and comparisons with other methods.
\subsection{Active + Passive Illumination} 
We first highlight the importance of having both active and passive illumination in our system. To show the impact of the two components, we run ASN on a typical living room scene with active only (i.e. all the lights in the room were off), passive only (laser from the sensor is off), and a combination of active and passive.  The results are shown in Fig. \ref{fig:actrive_passive}. The result with only active illumination is noisier (see the presence of dot spike) and loses object edges (e.g. the coffee table) due to the strong presence of the dot pattern. The result from traditional passive stereo matching fails in texture-less regions, such as the wall. In contrast, the combination of the two gives a smooth disparity with sharp edges. 


\subsection{Image Resolution}
\label{sec:resolution}
We here evaluate the importance of image resolution for active stereo. We hypothesize that high-resolution input images are needed to generate high-quality disparity maps because any downsizing would lose considerable high frequency signals from the active pattern.  To test this, we compare the performance of ASN trained with full and half resolution inputs in Fig. \ref{fig:qual_resolution}. Although deep architectures are usually good at super-resolution tasks, in this case there is a substantial degradation of the overall results. The thin structure and boundary details are missing in the ASN trained with half resolution.  Please note that this result drives our decision to design a compact network architecture to avoid excessive memory usage.

\begin{figure}[t]
    \centering
      \includegraphics[width=\columnwidth]{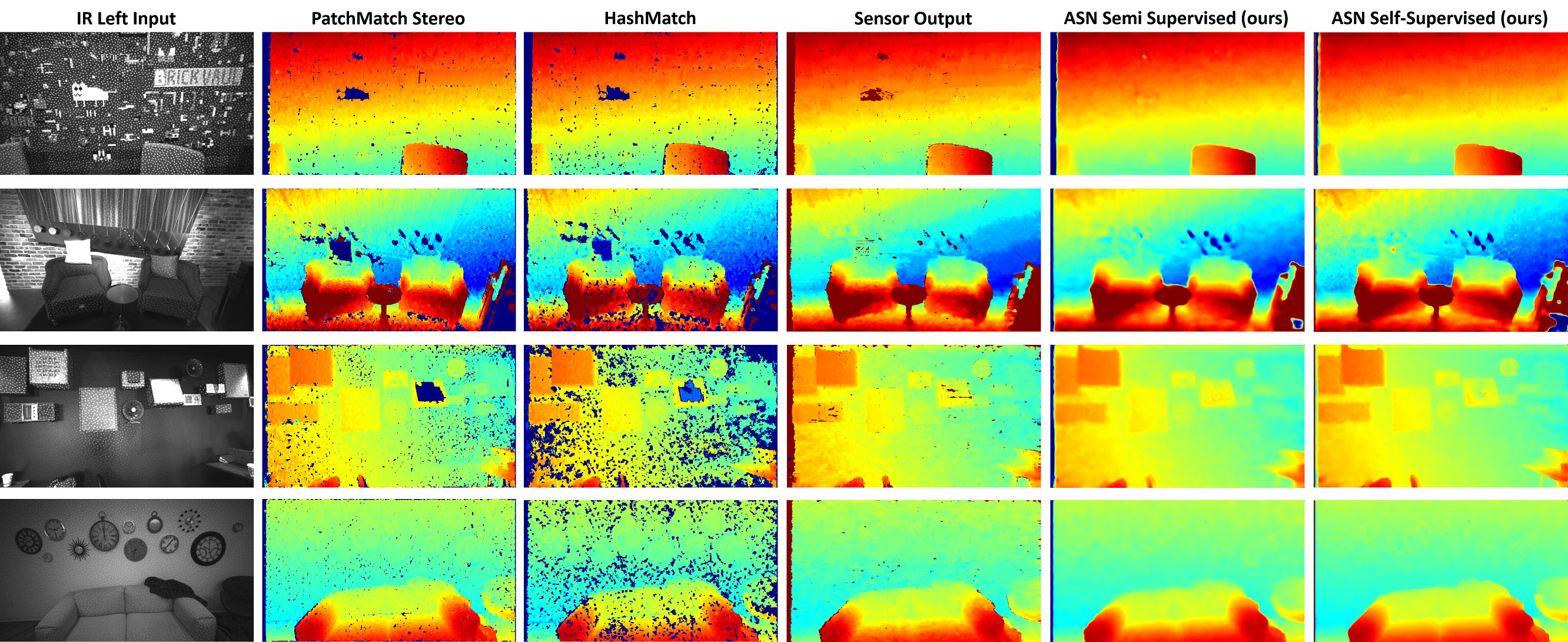}
    \caption{We provide more qualitative examples. Notice in these challenging scenes how we provide a more complete output, crisper edges and less oversmoothed disparities.}
 \label{fig:more_qual}
 \end{figure}

\begin{figure}[t]
    \centering
      \includegraphics[width=\columnwidth]{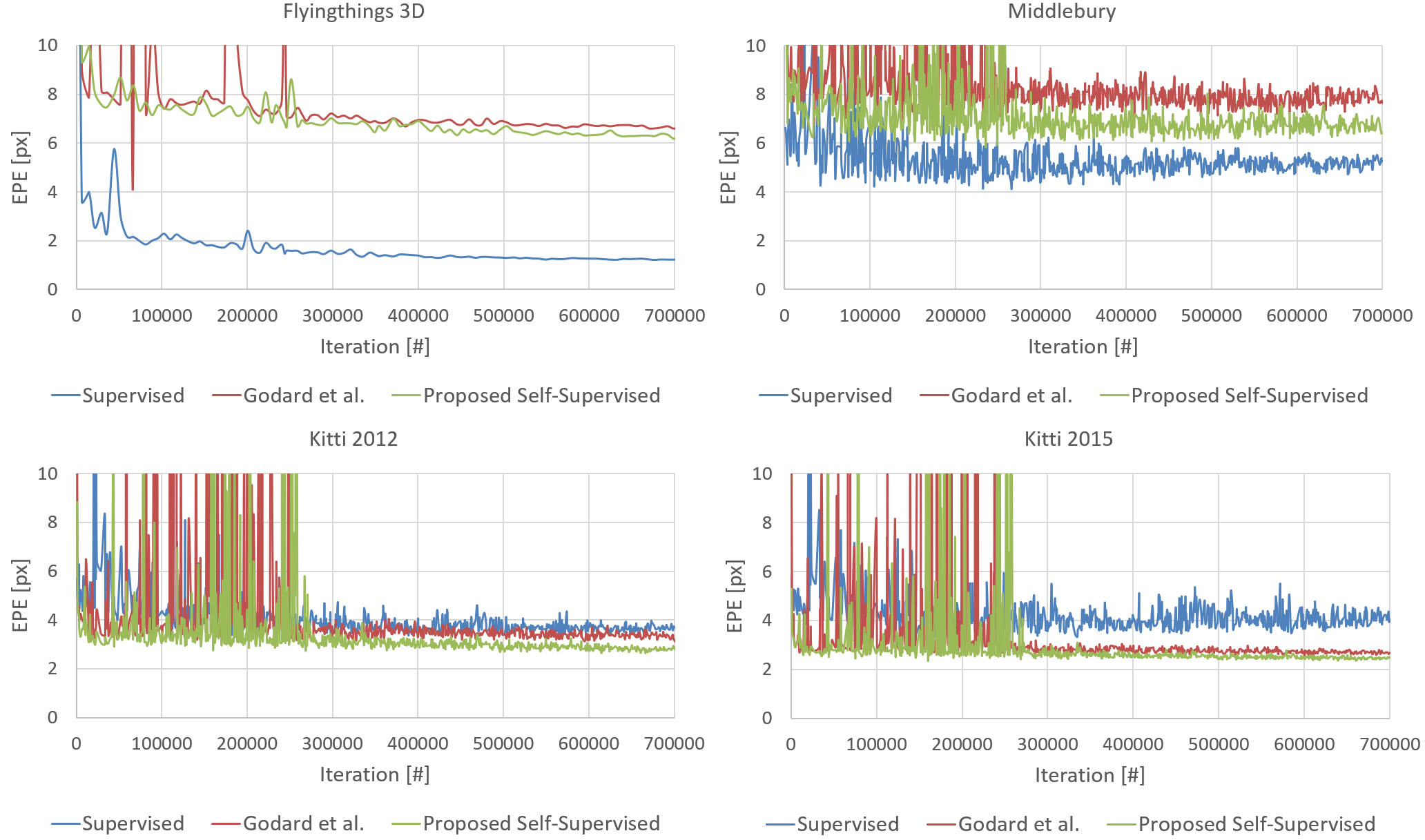}
    \caption{Quantitative Evaluation - Passive Stereo. Our method outperforms the state of the art unsupervised method by Godard et al. \cite{godard2017unsupervised} and it generalizes better than the supervised loss on the Kitti 2012 and Kitti 2015 datasets.}
 \label{fig:quantitative_rgb}
 \end{figure}

 \subsection{Performance w.r.t. dataset size}
\begin{figure*}[t]
\centering
\includegraphics[width=\linewidth]{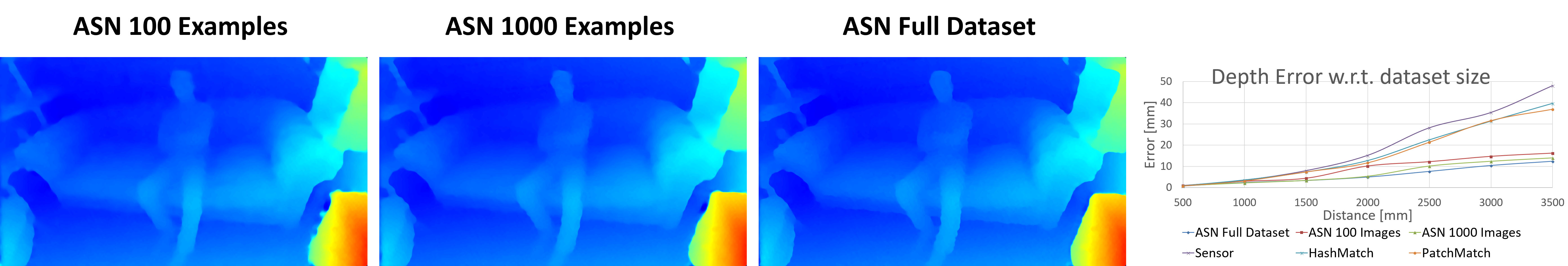}
\caption{ Analysis of performance w.r.t data size. We found the model to be effective even with a small dataset. $100$ images contain millions of individual pixels which are used as training examples. Increasing the dataset size leads to more accurate (sharper) depth maps, although the overall error does not improve that much. We did not observe any significant improvement beyond $10000$ examples.}
\label{fig:generalize}
\end{figure*}

 Here we show the performance of the algorithm with respect to the dataset size. In Fig. \ref{fig:generalize}, we show how our method is able to produce reasonable depthmaps on unseen images even when only a small dataset of $100$ images is used. Increasing the training data size, leads to sharper results. We did not find any additional improvement when more than $10000$ images are employed.

\subsection{More Qualitative Results}
\begin{figure}[t]
    \centering
      \includegraphics[width=\columnwidth]{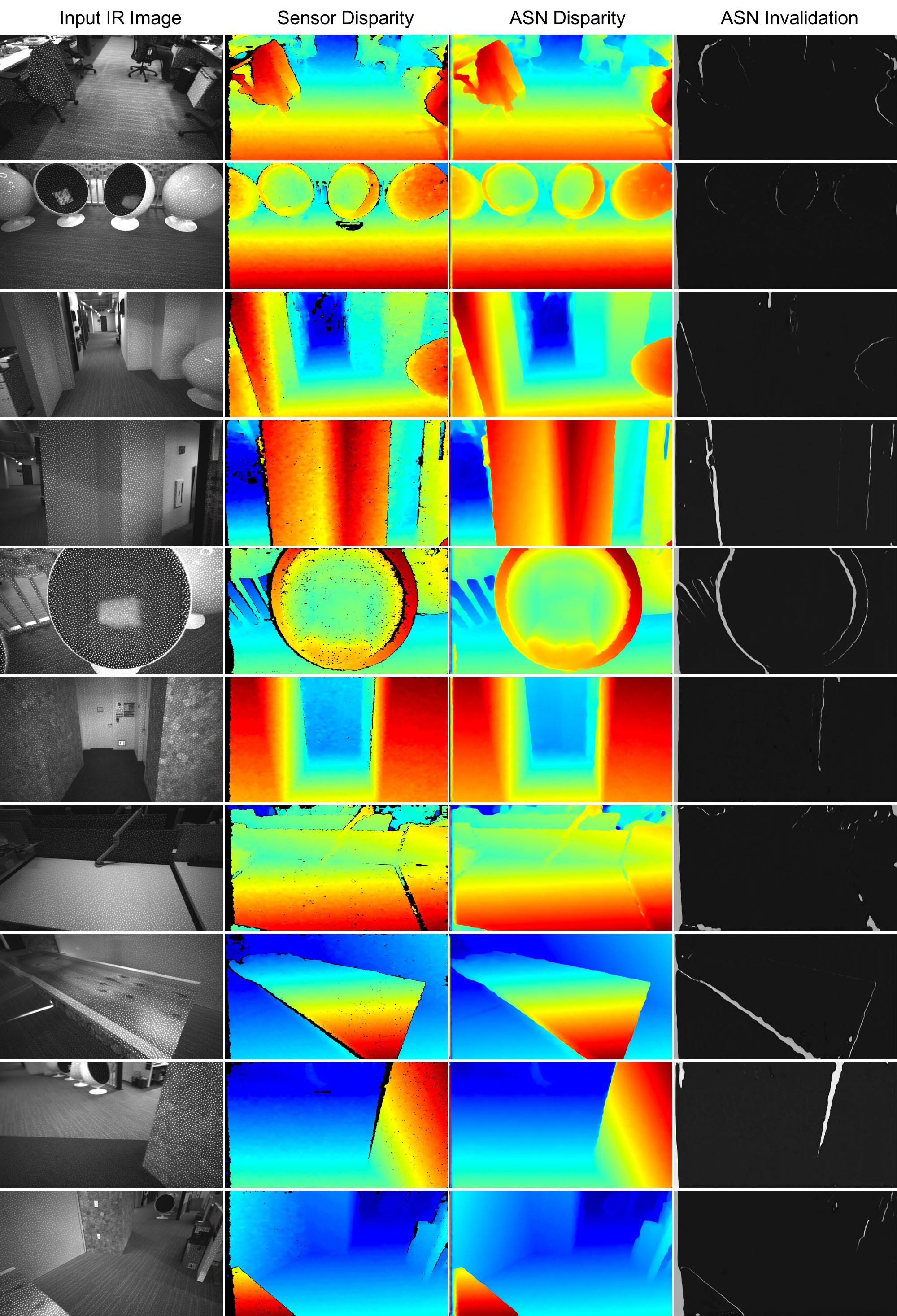}
    \caption{Results of estimated disparity and invalidation mask on more diverse scenes.}
 \label{fig:more_result}
 \end{figure}

Here we show additional challenging scenes and compare our method with local stereo matching algorithms as well as the sensor output. Results are depicted in Fig. \ref{fig:more_qual}. Notice how the proposed methods not only estimate more complete disparity maps, but suffer less from edge fattening and oversmoothing.
Fig. \ref{fig:more_result} shows more results of the predicted disparity and invalidation mask on more diverse scenes.
 
 \begin{figure}[t]
    \centering
      \includegraphics[width=\columnwidth]{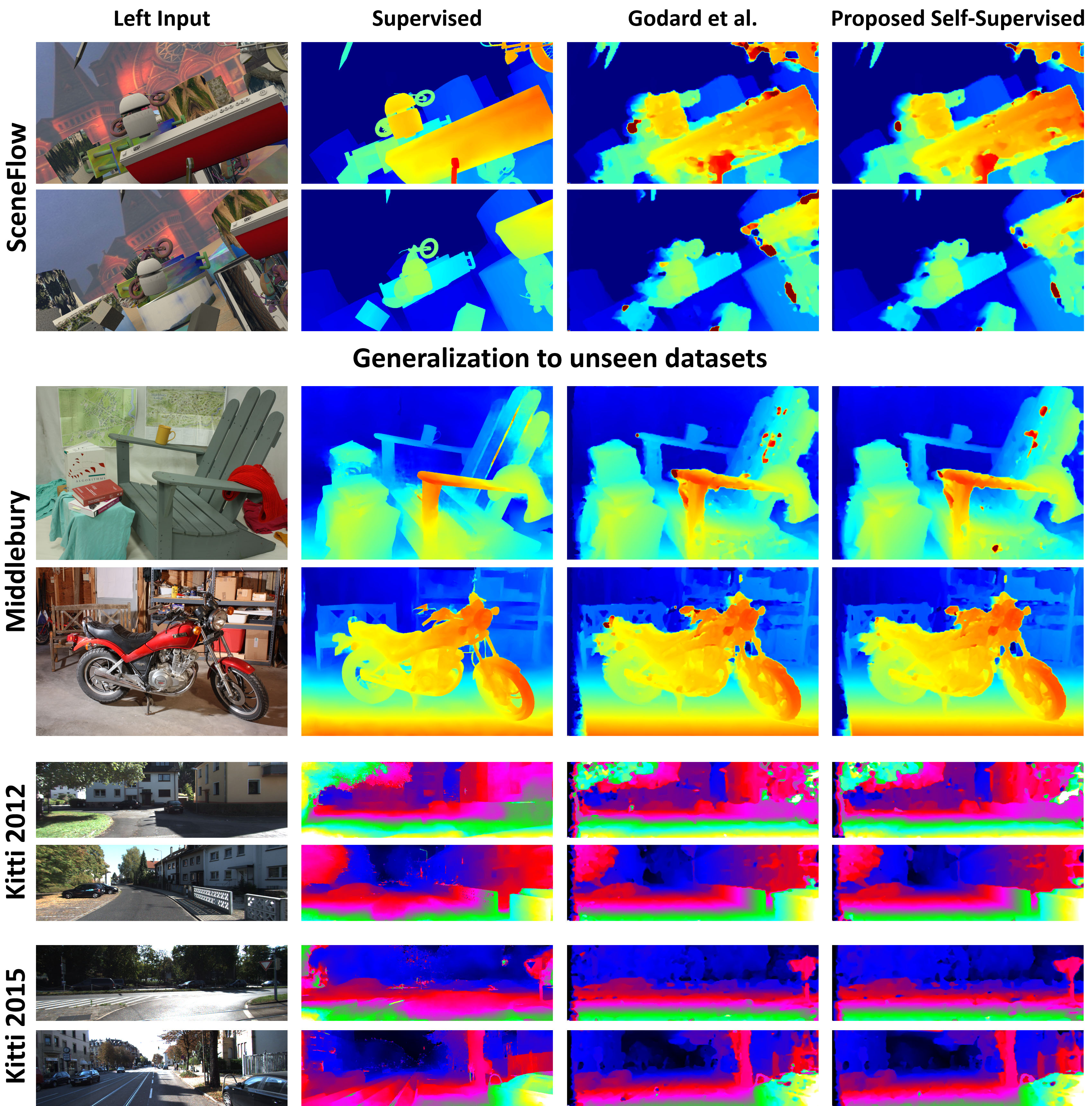}
    \caption{Examples of qualitative results on RGB data using our method. We trained on SceneFlow dataset and tested on the others without any finetuning. Notice that for Kitti dataset we show better results compared to the supervised methods which suffer from gross errors due to overfitting.}
 \label{fig:qual_rgb}
 \end{figure}

\section{Self-Supervised Passive Stereo}
We conclude our study with an evaluation of the proposed self-supervised training method on RGB images. Although this is out of the scope of this paper, we show how the proposed loss generalizes well also for passive stereo matching problems. We consider the same architecture described in Sec. \ref{sec:architecture2} and we train three networks from scratch on the SceneFlow dataset \cite{flyingthings} using three different losses: supervised, Godard et al. \cite{godard2017unsupervised} and our method. While training on SceneFlow, we evaluate the accuracy for the current iteration on the validation set as well as on unseen datasets such as Middlebury \cite{middlebury14}, Kitti 2012 and Kitti 2015 \cite{Geiger2012CVPR,Menze2015CVPR}. This is equivalent to testing our model on different datasets without fine-tuning. In Fig. \ref{fig:quantitative_rgb} we show the EPE error of all the trained methods.
Please note that the model trained using our loss achieves the lowest error on all datasets, which indicates that our method generalizes well. In particular on Kitti 2012 and Kitti 2015, we even outperform the supervised method trained on SceneFlow, which suggests the supervised loss is prone to overfitting. 

Finally, we conclude this work showing qualitative results on passive RGB images. Fig. \ref{fig:qual_rgb} we compare the three networks on the validation set used in SceneFlow (top two rows) as well as on unseen datasets (bottom 6 rows). Notice how our method always outperforms Godard et al. \cite{godard2017unsupervised} in terms of edge fattening and frequent outliers on every dataset. Moreover, on the Kitti images, we do not suffer from gross errors such as the supervised model, which clearly overfit the original dataset.


%



\end{document}